\documentclass[10pt,twocolumn,letterpaper]{article}

\usepackage{cvpr}      

\usepackage{graphicx}
\usepackage{amsmath}
\usepackage{amssymb}
\usepackage{booktabs}
\usepackage{subcaption}
\usepackage{multirow}
\usepackage{mmstyles}
\usepackage{bbm}
\usepackage[table]{xcolor}
\usepackage{colortbl}
\definecolor{myy}{RGB}{126,95,0}
\definecolor{mygray}{gray}{.9}
\definecolor{bblue}{RGB}{30,80,120}
\definecolor{mygray1}{gray}{.7}
\definecolor{ggray}{RGB}{127,127,127}
\definecolor{mygreen}{RGB}{93,174,86}
\usepackage[pagebackref,breaklinks,colorlinks]{hyperref}

\usepackage[capitalize]{cleveref}
\crefname{section}{Sec.}{Secs.}
\Crefname{section}{Section}{Sections}
\Crefname{table}{Table}{Tables}
\crefname{table}{Tab.}{Tabs.}

\def\cvprPaperID{2990} 
\def\confName{CVPR}
\def\confYear{2022}

\begin{document}

\title{Cross Language Image Matching for Weakly Supervised Semantic Segmentation}

\author{Jinheng Xie$^{\dag}$, Xianxu Hou$^{\dag}$, Kai Ye, Linlin Shen\thanks{Corresponding Author}\\
	School of Computer Science \& Software Engineering, Shenzhen University, China\\
	Shenzhen Institute of Artificial Intelligence of Robotics of Society, Shenzhen, China\\
	Guangdong Key Laboratory of Intelligent Information Processing, Shenzhen University, China\\
	National Engineering Laboratory for Big Data System Computing Technology, Shenzhen University, China\\
	{\tt\small \{xiejinheng,yekai\}2020@email.szu.edu.cn, hxianxu@gmail.com, llshen@szu.edu.cn}
}
\maketitle
\def\thefootnote{$^{\dag}$}\footnotetext{Equal Contribution}
\begin{abstract}
It has been widely known that CAM (Class Activation Map) usually only activates discriminative object regions and falsely includes lots of object-related backgrounds. As only a fixed set of image-level object labels are available to the WSSS (weakly supervised semantic segmentation) model, it could be very difficult to suppress those diverse background regions consisting of open set objects. In this paper, we propose a novel Cross Language Image Matching (CLIMS) framework, based on the recently introduced Contrastive Language-Image Pre-training (CLIP) model, for WSSS. The core idea of our framework is to introduce natural language supervision to activate more complete object regions and suppress closely-related open background regions. In particular, we design object, background region and text label matching losses to guide the model to excite more reasonable object regions for CAM of each category. In addition, we design a co-occurring background suppression loss to prevent the model from activating closely-related background regions, with a predefined set of class-related background text descriptions. These designs enable the proposed CLIMS to generate a more complete and compact activation map for the target objects. Extensive experiments on PASCAL VOC2012 dataset show that our CLIMS significantly outperforms the previous state-of-the-art methods. Code will be available at \href{https://github.com/CVI-SZU/CLIMS}{https://github.com/CVI-SZU/CLIMS}.
\end{abstract}

\section{Introduction}
\label{sec:intro}
\begin{figure}[t]
	\centering
	\includegraphics[width=\linewidth]{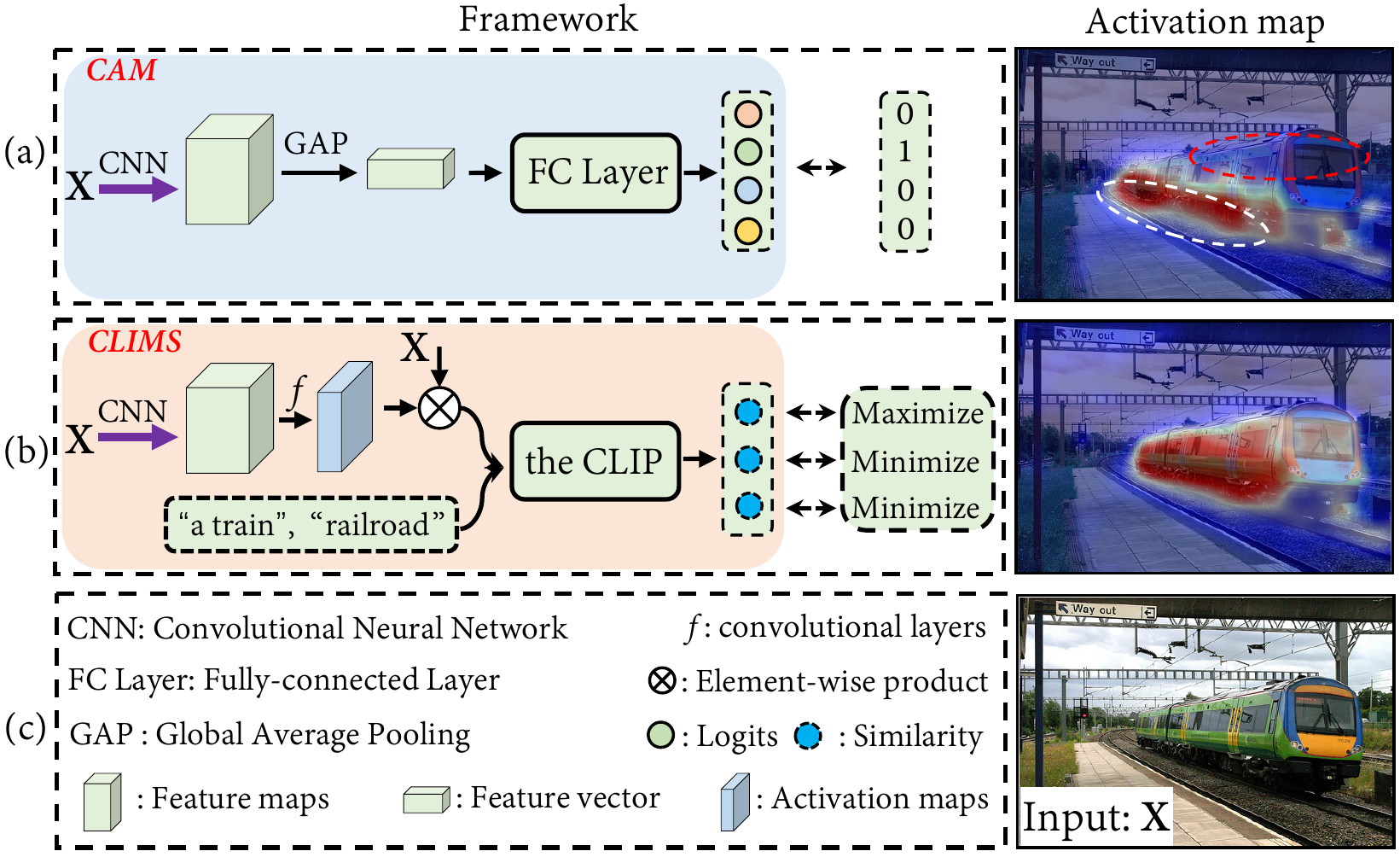}
	\vspace{-16pt}
	\caption{(a) Conventional CAM solution. (b) The proposed CLIMS. The problem of false-activation of irrelevant background, e.g., railroad and ground, and underestimation of object contents usually exist in conventional CAM method. To solve this problem, we propose a novel text-driven learning framework, CLIMS, which introduces natural language supervision, i.e., an open-world setting, for exploring complete object contents and excluding irrelevant background regions. Best viewed in color.}  
	
	\label{fig:motivation}
	\vspace{-14pt}
\end{figure}

Semantic segmentation attempts to assign each pixel in an image with a semantic label. Though fully supervised semantic segmentation has achieved remarkable success in recent years, pixel-level annotation is significantly time-consuming and labor-intensive. Instead, weakly supervised semantic segmentation (WSSS) tries to mitigate this issue by relying solely on image-level~\cite{affinitynet, sec, constrained, pedro}, bounding box-level~\cite{wsemi, boxsup}, point-level~\cite{whatpoint}, or scribble-based supervision~\cite{scribblesup, rw}. This work aims to only use image-level labels in the learning of a semantic segmentation model.

Existing WSSS approaches typically follow a three-stage learning process. First, the image-level labels are used as supervision at the feature level to train a classification network to generate initial class activation maps (CAMs) (as shown in the left of Fig.~\ref{fig:motivation}(a)). Then the initial CAMs are refined as pseudo ground-truth masks using dense CRF~\cite{dcrf}, pixel affinity-based methods~\cite{affinitynet, irnet}, or additional saliency maps~\cite{oaa, liid, eps}. Finally, the refined pseudo ground-truth masks are used to further train a segmentation network. However, as only a fixed set of object categories is available during the first stage training, i.e., a close-world setting, class-related background pixels, e.g., railroad, also contribute to the prediction of closely related objects, e.g., train. This results in unnecessary activation of background in the generation of initial CAMs, as shown in the right of Fig.~\ref{fig:motivation}(a). Besides, conventional CAM solution usually struggles in the underestimation of object contents. Both of them severely limit the quality of initial CAMs for the next two stages.

In this paper, we design a novel Cross Language Image Matching framework for WSSS, i.e., CLIMS, based on the power of recently introduced Contrastive Language-Image Pre-training (CLIP)~\cite{clip} model, to address the aforementioned issues. The CLIP model is pretrained from scratch on a dataset of 400 million image-text pairs (automatically collected from the publicly available sources on the Internet), which enables CLIP to associate much wider visual concepts in the image with their text labels in an open-world setting, rather than a fixed set of predetermined object categories. Based on this, the proposed CLIMS has great potentials to generate a high-quality initial activation map for each object category without irrelevant background (as shown in the right of Fig.~\ref{fig:motivation}(c)). 
	
In the left of Fig.~\ref{fig:motivation}(a), the conventional CAM method performs image-level supervision on the average feature after the global average pooling (GAP) layer. Given the trained model, the class activation maps (CAMs) can be extracted. However, the CLIP model cannot be directly used in this pipeline. Instead, as shown in Fig.~\ref{fig:motivation}(b), we replace the GAP and fully connected (FC) layer with convolutional layers to directly generate an activation map for each class under the supervision from CLIP model, where natural language can be used to guide the model for the activation maps generation. 

\begin{figure*}
	\centering
	\includegraphics[width=\linewidth]{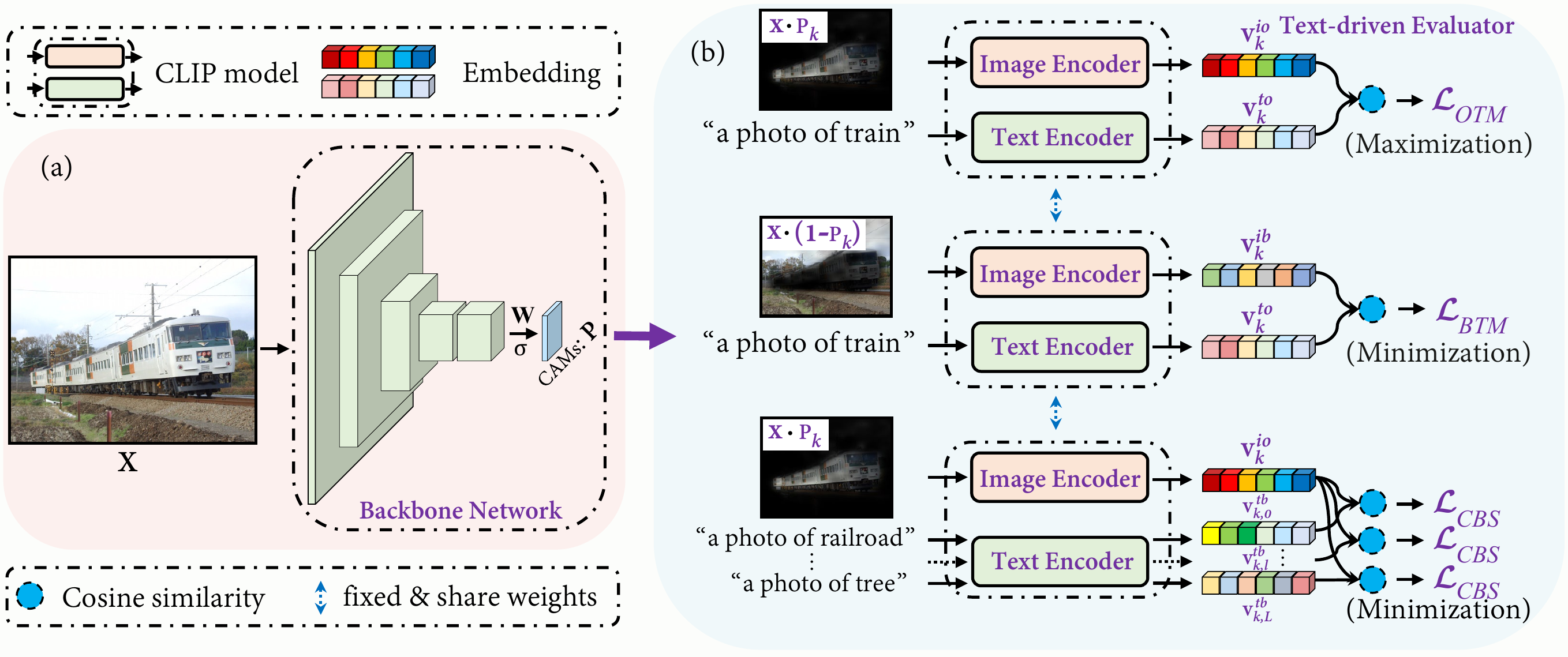}
	\caption{An overview of the proposed Cross Language Image Matching framework for WSSS, i.e., CLIMS. (a) The backbone network for predicting initial CAMs. $\sigma$ denotes the sigmoid activation function. $\mW$ denotes the weight matrix of convolutional layers. (b) The text-driven evaluator. It consists of three CLIP-based loss functions, i.e., object region and text label matching loss $\mathcal{L}_{OTM}$, background region and text label matching loss $\mathcal{L}_{BTM}$, and co-occurring background suppression loss $\mathcal{L}_{CBS}$. Best viewed in color.}
	\label{fig:overall_framework}
	\vspace{-10pt}
\end{figure*}

Details of the proposed CLIMS are depicted in Fig.~\ref{fig:overall_framework}. It mainly consists of a backbone network and a text-driven evaluator including three CLIP-based loss functions, i.e., Object region and Text label Matching loss ($\mathcal{L}_{OTM}$), Background region and Text label Matching loss ($\mathcal{L}_{BTM}$), and Co-occurring Background Suppression loss ($\mathcal{L}_{CBS}$). The core idea is to complementarily learn the generation of initial CAMs $\mP$ through the supervision of the text-driven evaluator. First, given an image, the backbone network predicts the initial CAMs $\mP$, which represents the probability of each pixel belonging to a category. $\mP$ and $(1-\mP)$ are then multiplied with the input image to mask out the object and background area, respectively, which serve as the input of the text-driven evaluator. As shown in Fig.~\ref{fig:overall_framework}(b), each mask-out area and its corresponding text category label are passed to the CLIP model to calculate their cosine similarity. The text labels for foreground objects,  e.g., “train”, “cat” and “person”, etc, are defined according to the dataset. During training, $\mathcal{L}_{OTM}$ aims to maximize the similarity between the foreground object area and the given text label, e.g., “a photo of train”. In this way, though the generated CAMs can gradually approach the target object in the image, the completeness of the regions is not guaranteed. For example, the image still looks like a bird even when only head of the bird is visible. Thus, we propose $\mathcal{L}_{BTM}$ to minimize the similarity between mask-out foreground area $\mX \cdot (1-\mP)$ and the same object category text label as $\mathcal{L}_{OTM}$. This excludes object regions out of $(1-\mP)$ and recovers more probable object contents in $\mP$. In addition, to constrain the size of activated regions, we design a regularization term to ensure the compactness of $\mP$. However, when the object area is activated, the background closely related to the object, e.g., train and railroad, boat and river, etc, will usually be activated as well, as no pixel-level labels are available. To address this issue, we additionally define a set of class-related background text labels, such as “railroad” (background of the train) and “river” (background of the boat), etc. Based on these text labels and the CLIP model, we design $\mathcal{L}_{CBS}$ to minimize the similarity between the mask-out object area $\mX \cdot \mP$ and these co-occurring background text labels. This enables CLIMS to exclude the irrelevant class-related background, e.g., railroad, out of the initial CAMs $\mP$. These CLIP-based loss functions work complementarily to refine the generated initial CAMs by the exploration of more complete object areas and exclusion of irrelevant background pixels.

Collectively, the main contributions of this paper can be summarized as:
\begin{itemize}
	\item We propose a text-driven learning framework, CLIMS, to introduce image-text matching model based supervision, i.e., an open-world setting, for WSSS.
	\item We design three CLIP-based loss functions and an area regularization. The object, background region and matching losses ensure the correctness and completeness of initial CAMs. The co-occurring background suppression loss can further substantially mitigate the influence of class-related background. The area regularization can constrain the size of activated regions.
	
	\item Extensive experiments on PASCAL VOC2012 dataset show that the proposed CLIMS significantly outperforms the previous state-of-the-art methods.
	
\end{itemize}

\section{Related Work}
In this section, we will first review existing weakly supervised semantic segmentation methods according to the three-stage learning process of WSSS: initial CAMs generation, CAMs refinement techniques, and segmentation network training. In addition, we will briefly discuss the CLIP model, which serves as the motivation for this work.

\textbf{Weakly supervised semantic segmentation.} The pipeline of the conventional CAM~\cite{zhou2016learning} solution has been used in the majority of previous WSSS works. Hou \etal \cite{seenet} propose two self-erasing strategies for focusing attention only on the reliable regions, generating complete initial CAMs. Chang \etal \cite{sc-cam} propose to investigate object sub-categories to mine more object parts and then improve the completeness of initial CAMs. Sun \etal \cite{mcis} incorporate two neural co-attentions into the classifier for discovering shared or unshared semantics in a pair of training images. This helps to extract more complete initial CAMs from the classifier. Jungbeom \etal~\cite{advcam} propose an anti-adversarial manner to discover more regions of the target object in the activation map. Ahn and Kwak \cite{affinitynet} design a deep neural network, called AffinityNet, to predict semantic affinity between a pair of adjacent image coordinates. This semantic affinity is then applied to refine the generated initial CAMs as pseudo ground-truth masks. Previous works~\cite{oaa, liid, edam, eps} instead use additional saliency maps from a fully supervised saliency detector to refine the generated initial CAMs. The series of DeepLab~\cite{deeplabv1, deeplabv2} models are typically used to train a semantic segmentation network with the pseudo ground-truth masks. 

\textbf{Contrastive Language-Image Pre-training (CLIP).} The contrastive language-image pretraining (CLIP)~\cite{clip} shows great success and potential on many vision tasks in a zero-shot setting. The CLIP model consists of an image encoder and a text encoder. Given a batch of image and text pairs, the CLIP model learns the embedding to measure the similarity between image and text. The CLIP model is trained on a large dataset of 400 million image-text pairs, the set of object categories CLIP can recognize is much bigger and diverse than a fixed set of object categories in a small dataset, e.g., PASCAL VOC2012~\cite{pascal-voc-2012}. The image-text pairs are automatically collected from the Internet, without the involvement of manual efforts. 
\section{Methodology}
\label{sec:method}
In this section, we will first review the conventional CAM pipeline and its limitations. The proposed learning framework, i.e., CLIMS, is then introduced. Finally, we will detail three CLIP-based loss functions: object region and text label matching loss, background region and text label matching loss and co-occurring background suppression loss, and an area regularization.
\subsection{Revisiting the Conventional CAM}
The majority of previous WSSS works follow the CAM pipeline to generate an initial activation map for each category in the image. Given an input image $\mX$ and corresponding image-level label $y \in \mathbb{R}^{1\times K}$, a backbone network will first embed $\mX$ to high-level feature maps $\mZ \in \mathbb{R}^{C\times H\times W}$, where $K$ denotes the number of classes, $C$ and $H\times W$ denote the number of channels and spatial dimension, respectively. A global average pooling (GAP) layer and a $1\times 1$ convolution layer with learnable matrix $\mW \in \mathbb{R}^{C\times K}$ are then applied on $\mZ$ to produce prediction logits $\hat{y} \in \mathbb{R}^{1\times K}$. During training, the sigmoid cross entropy loss is calculated as follow:
\begin{equation}
	\mathcal{L}(\hat{y}, y)=-\sum^{K}_{k=1} y_k \cdot \log \sigma\left(\hat{y_k}\right)+\left(1-y_k\right) \cdot \log \left(1-\sigma\left(\hat{y_k}\right)\right),
\end{equation}
where $\sigma$ is the sigmoid activation function.

Given the trained backbone network, $\mW$ is directly applied on $\mZ$ to generate initial CAMs $\mP \in \mathbb{R}^{K\times H\times W}$:
\begin{equation}
	\mP_k(h,w) = \mW_k^\top\mZ(h,w),
\end{equation}
where $\mZ(h,w)$ denotes the representation vector located on $(h,w)$. The corresponding weight vector and activation map for a given object category is $\mW_k$ and $\mP_k$. Due to the limited supervision, conventional CAM, despite being simple and effective, may struggle with the only activation of discriminative object parts and the unnecessary activation of closely related backgrounds.
\subsection{Cross Language Image Matching Framework}
Fig.~\ref{fig:overall_framework} depicts an overview of CLIMS. The backbone network illustrated in Fig.~\ref{fig:overall_framework}(a) is similar to conventional CAM solution, except that the GAP layer is removed and a sigmoid function $\sigma$ is directly applied after $\mW$: 
\begin{equation}
	\mP_k(h,w)= \sigma(\mW_k^\top\mZ(h,w)).
\end{equation}

While conventional WSSS methods only use the supervision of a fixed set of predetermined object categories, we propose the text-driven evaluator based on the CLIP model to explore additional object categories in the dataset. As shown in Fig.~\ref{fig:overall_framework}(b), the text-driven evaluator consists of an image encoder $f_i(\cdot)$ and a text encoder $f_t(\cdot)$ from the CLIP model. To begin with, $\mP_k$ and $(1-\mP_k)$ are multiplied by $\mX$ to mask out the foreground object and background pixels, respectively. The results are then mapped to representation vectors $\vv_k^{io}$ and $\vv_k^{ib}$ by $f_i(\cdot)$:
\begin{equation}
	\vv_k^{io} = f_i(\mX \cdot \mP_k ),\quad \vv_k^{ib} = f_i(\mX \cdot  (1-\mP_k) ).
\end{equation}

Following CLIP~\cite{clip}, the corresponding object text prompt $\vt^o_k$ for $(\mX \cdot \mP_k )$ is represented as “a photo of $\{ \}$”, e.g., “a photo of the train”. On the contrary, the corresponding class-related background text prompts $\vt^b_{k,l}$ are manually pre-defined as a set of $L$ co-occurring background closely related with the object of the $k$-th category. For example, the class-related background of boat ($k$-th object) is $\{$“a photo of river”, “a photo of a lake”$\}$. Then $\vt_{k,0}^b$=$\{$“a photo of river”$\}$, $\vt_{k,1}^b$=$\{$“a photo of a lake”$\}$, the text representations can be obtained as follow:
\begin{equation}
	\vv_k^{to} = f_t(\vt_k^o),\quad \vv_{k,l}^{tb} = f_t(\vt_{k,l}^b),
\end{equation}
where $\vt_k^o$ and $\vt_{k,l}^b$ denote text label of the object and the $l$-th class-related co-occurring background for the specific class $k$, respectively.

\subsection{Object region and Text label Matching}
Given the $k$-th foreground object representation $\vv^{io}_k$ and its corresponding text representation $\vv^{to}_k$, we begin by calculating the cosine similarity between image and text representations and then maximize it using the proposed object region and text label matching loss $\mathcal{L}_{OTM}$:
\begin{equation}
	\mathcal{L}_{OTM} = -\sum_{k=1}^{K} y_k \cdot \log (s^{oo}_k), 
\end{equation}
\begin{equation}
 s^{oo}_k=\text{sim}(\vv^{io}_k, \vv^{to}_k),
\end{equation}
$s^{oo}_k$ indicates the cosine similarity between $\vv^{io}_k$ and $\vv^{to}_k$. The generated initial CAMs will gradually approach the target object under the supervision of $\mathcal{L}_{OTM}$. However, the $\mathcal{L}_{OTM}$ alone can not encourage the model to explore non-discriminative object regions and suppress the background regions activated in $\mP_k$.   
\subsection{Background region and Text label Matching}
To improve the completeness of activated object regions, we design the background region and text label matching loss $\mathcal{L}_{BTM}$ to include more object contents. Given the background representation $\vv^{ib}_k$ and its corresponding text representation $\vv^{to}_k$ (note that, the text label for $\mathcal{L}_{BTM}$ is the same with that for $\mathcal{L}_{OTM}$), the $\mathcal{L}_{BTM}$ is calculated as follow:
\begin{equation}
	\mathcal{L}_{BTM} = -\sum_{k=1}^{K} y_k \cdot \log (1 - s^{bo}_k),
\end{equation}
\begin{equation}
 s^{bo}_k=\text{sim}(\vv^{ib}_k, \vv^{to}_k),
\end{equation}
where $s^{bo}_k$ represents the cosine similarity between $\vv^{ib}_k$ and $\vv^{to}_k$. When $\mathcal{L}_{BTM}$ is minimized, fewer target object pixels are reserved in $\mX \cdot (1-\mP_k)$ and more target object contents are recovered in $(\mX \cdot \mP_k)$. This ensures that more complete object contents are activated in $\mP_k$.
\subsection{Co-occurring Background Suppression}
The aforementioned two loss functions, however, only ensure $\mP$ to completely cover the target object, without taking into account the false-activation of co-occurring class-related background. The inclusion of co-occurring background may significantly degrade the quality of generated pseudo ground-truth masks. However, pixel-level labeling of these backgrounds is very time-consuming and labor expensive, and is usually not available in WSSS. As the set of background is much more diverse than that of foreground, lots of them might not have been seen by the classification network trained using ImageNet. However, it's much easier to use the pretrained CLIP to identify these backgrounds given the corresponding text descriptions. To solve this problem, we design the following co-occurring background suppression loss. Given the target object representation $\vv^{io}_k$ and its corresponding text representation of class-related background $\vv^{tb}_{k,l}$, the loss is calculated as: 
\begin{equation}
	\mathcal{L}_{CBS} = -\sum_{k=1}^{K}\sum_{l=1}^{L} y_k \cdot \log (1 - s^{ob}_{k,l}),
\end{equation}
\begin{equation}
	 s^{ob}_{k,l}=\text{sim}(\vv^{io}_{k}, \vv^{tb}_{k,l}),
\end{equation}
where $s^{ob}_{k,l}$ indicates the cosine similarity between $\vv^{io}_k$ and $\vv^{tb}_{k,l}$. During training, the backbone network will gradually suppress the false-activation of class-related background regions in $\mP_k$ for the minimization of $\mathcal{L}_{CBS}$.

\subsection{Area Regularization}
With only $\mathcal{L}_{OTM}$, $\mathcal{L}_{BTM}$, and $\mathcal{L}_{CBS}$, if both the irrelevant backgrounds and the target object are included in the activation map, the CLIP model can still correctly predict the target object. Therefore, we design a pixel-level area regularization term to constraint the size of activation maps to ensure that the irrelevant backgrounds are excluded in the activation map $\mP_k$:
\begin{equation}
	\label{eq:area_loss}
	\mathcal{L}_{REG} = \frac{1}{K}\sum_{k=1}^{K} S_k,\quad  \textrm{where}\quad S_{k} =  \frac{1}{HW}  \sum_{h=1}^{H}\sum_{w=1}^{W} \mP_{k}(h,w).
\end{equation}

\subsection{Overall Training Objective}
The overall training loss for the proposed text-driven learning framework CLIMS can be formulated as:
\begin{equation}
	\label{eq:13}
	\mathcal{L} = \alpha \mathcal{L}_{OTM} + \beta \mathcal{L}_{BTM} + \gamma \mathcal{L}_{CBS} + \delta \mathcal{L}_{REG},
\end{equation}
where $\alpha$, $\beta$, $\gamma$, and $\delta$ are the hyper-parameters weighting the four loss terms.

\begin{figure*}[!htbp]
	\centering
	\includegraphics[width=\linewidth]{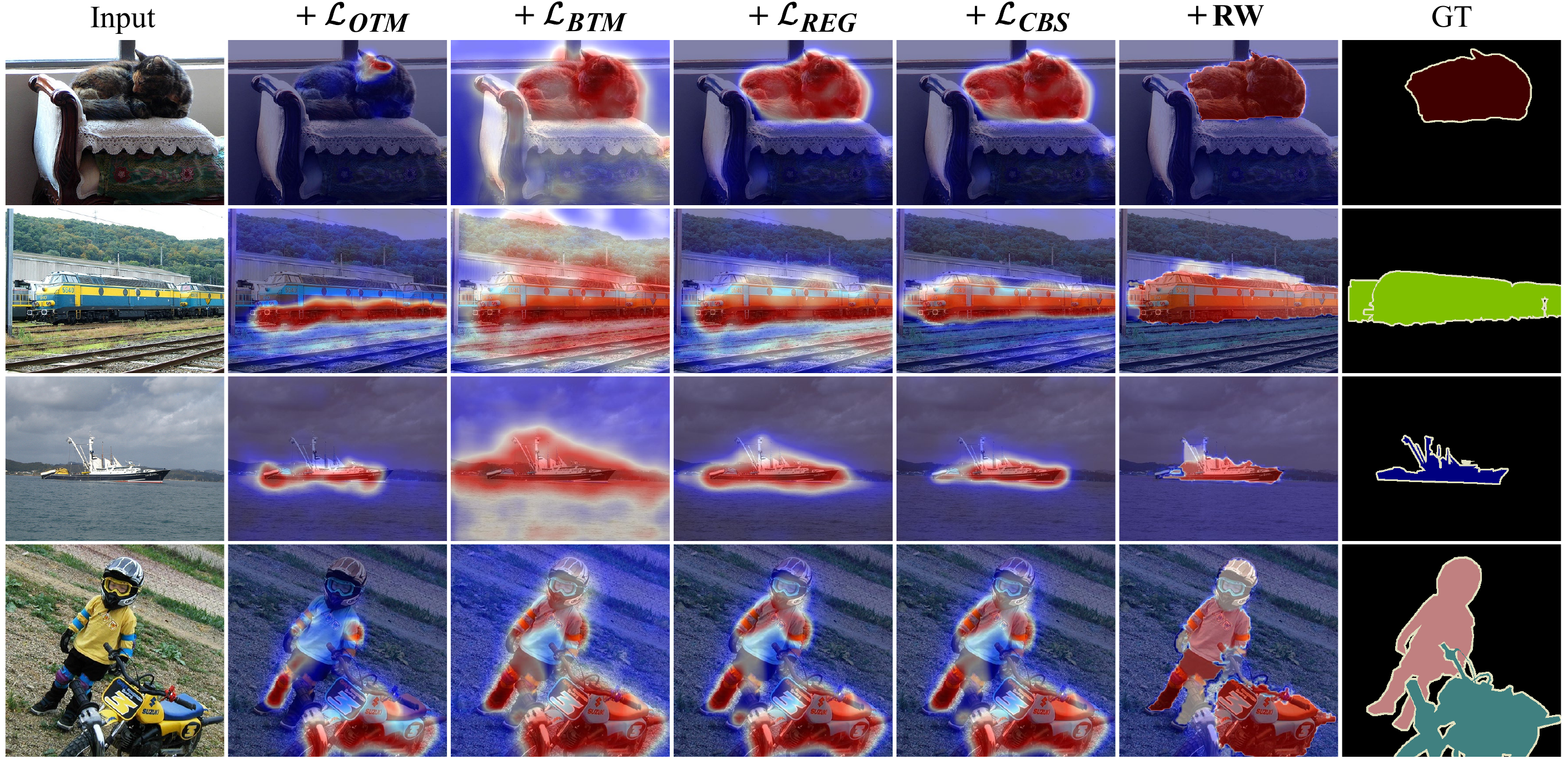}
	\vspace{-20pt}
	\caption{Initial CAMs generated by the proposed CLIMS using different combinations of loss functions. Input images are shown in column 1. Columns 2 to 5 present the generated CAMs using $\mathcal{L}_{OTM}$, $\mathcal{L}_{OTM}+\mathcal{L}_{BTM}$, $\mathcal{L}_{OTM}+\mathcal{L}_{BTM}+\mathcal{L}_{REG}$, and $\mathcal{L}_{OTM}+\mathcal{L}_{BTM}+\mathcal{L}_{REG}+\mathcal{L}_{CBS}$, respectively. \textbf{RW} denotes the refinement of PSA~\cite{affinitynet}. Best viewed in color.}
	\label{fig:ablation}
	\vspace{-10pt}
\end{figure*}

\section{Experiments}
\subsection{Experimental Setup}
\textbf{Datasets and Evaluation Metric.} PASCAL VOC2012~\cite{pascal-voc-2012} is a popular semantic segmentation dataset with 20 object categories, which consists of 1,464 images for training, 1,449 images for validation, and 1,456 images for test. Following the common protocol in previous studies, we train the proposed CLIMS using an augmented training set of 10,582 images. The mean Intersection over Union (mIoU) is adopted as the evaluation metric for all experiments.

\textbf{Implementation Details.} The input image is randomly rescaled and then augmented by random cropping to $512\times 512$. Horizontal flipping is also used for augmenting training data. SGD is adopted as the default optimizer. The cosine annealing policy is applied to schedule learning rate. The default batch size is 16. As there are minor differences in describing classes between PASCAL VOC2012 and CLIP, we use the text label descriptions in the training set to finetune the CLIP model (both image and text encoder) for 20 epochs, with an initial learning rate of 0.00005 and a weight decay of 0.003. The CLIMS model is trained for 10 epochs, with an initial learning rate of 0.00025 and a weight decay of 0.0001. We follow ~\cite{affinitynet} to adopt ResNet-50~\cite{resnet} as backbone network for the generation of initial CAMs. All models are implemented in PyTorch and trained on NVIDIA A100 GPU with 40 GB memory.

\textbf{Refinement of initial CAMs.} Because the initial CAMs only coarsely cover the target object, refinement techniques such as PSA~\cite{affinitynet} and IRNet~\cite{irnet} are commonly used to improve the quality of initial CAMs before using them as pseudo ground-truth masks. To make a fair comparison, we follow SEAM~\cite{seam}, PuzzleCAM~\cite{puzzlecam}, and AdvCAM~\cite{advcam} to adopt PSA~\cite{affinitynet} for initial CAM refinement. 
\begin{table}[t]
	\centering
	
	\caption{Comparison of the quality of initial CAMs and refined pseudo ground-truth masks using \textbf{RW} (PSA~\cite{affinitynet}) on PASCAL VOC2012. The mIoU values here are reported on the \textit{train} set. Bac. denotes the backbone network for CAMs generation.}
	\label{tab:refine}
	\vspace{-10pt}
	\resizebox{\linewidth}{!}{
		\begin{tabular}{l c c c}
			\toprule
			Method & Bac. & CAMs & +\textbf{RW}\\
			\midrule  
			PSA$_\text{ CVPR'2018}$~\cite{affinitynet}   & WR38 & 48.0 & 61.0 \\
			SC-CAM$_\text{ CVPR'2020}$~\cite{sc-cam}    & WR38 & 50.9 & 63.4 \\
			SEAM$_\text{ CVPR'2020}$~\cite{seam}    & WR38& 55.4 & 63.6 \\
			PuzzleCAM$_\text{ ICIP'2021}$~\cite{puzzlecam}    & R50 & 51.5 & 64.7 \\
			VWE$_\text{ IJCAI'2021}$~\cite{vwe}    & R50 & 52.9 & - \\
			AdvCAM$_\text{ CVPR'2021}$~\cite{advcam}    & R50& 55.6 & 68.0 \\
			\rowcolor{mygray}
			CLIMS (Ours)    & R50 & \textbf{56.6}  & \textbf{70.5} \\
			\bottomrule
	\end{tabular}}
	\vspace{-10pt}
\end{table}

\begin{table}[h]
	\centering
	\caption{Evaluation results on PASCAL VOC2012 \textit{val} and \textit{test} sets. The best results are in \textbf{bold}. Sup. denotes the weak supervision type. $\mathcal{F}$ denotes full supervision. $\mathcal{S}$ denotes saliency map supervision. $\mathcal{I}$ denotes image-level supervision. Seg. denotes the segmentation network. Bac. denotes the backbone network for CAMs generation. V1: DeepLabV1. V2: DeepLabV2. V16: VGG-16~\cite{vgg}. R50: ResNet-50~\cite{resnet}. WR38: WideResNet38~\cite{resnet38}. $^\ddag$: Segmentation network pretrained using MS COCO dataset.}
	\vspace{-10pt}
	\resizebox{\linewidth}{!}{
		\begin{tabular}{c l l c c c}
			\toprule
			Sup. &Method & Seg. & Bac. & \textit{val} & \textit{test}\\
			\midrule
			\multirow{4}{*}{$\mathcal{F}$}& \textit{\textbf{Full supervision.}}\\
			&DeepLabV1$_\text{ ICLR'15}$~\cite{deeplabv1}  &  -  & - &  75.5 &  - \\
			&DeepLabV2$_\text{ TPAMI'18}$~\cite{deeplabv2}  &  -  & - & 77.6 &  79.7 \\
			&WideResNet38$_\text{ PR'19}$~\cite{resnet38}  &  -  & - &  80.8 &  82.5 \\
			\midrule
			\multirow{9}{*}{$\mathcal{I+S}$}& \multicolumn{5}{l}{\textit{\textbf{Image-level supervision + Saliency maps.}}}\\
			&OAA+$_\text{ ICCV'19}$~\cite{oaa} & V1$^\ddag$ & - & 65.2 & 66.4 \\
			&MCIS$_\text{ ECCV'20}$~\cite{mcis} & V1$^\ddag$ & V16 & 66.2 & 66.9 \\
			&LIID$_\text{ TPAMI'21}$~\cite{liid} & V2 & R50 & 66.5 & 67.5 \\
			&NSROM$_\text{ CVPR'21}$~\cite{nsrom} & V2$^\ddag$ & V16 & 68.3 & 68.5 \\
			&DRS$_\text{ AAAI'21}$~\cite{drs} & V2$^\ddag$ & V16 & 70.4 & 70.7 \\
			&EPS$_\text{ CVPR'21}$~\cite{eps} & V2$^\ddag$ & WR38 & 70.9 & 70.8 \\
			&EDAM$_\text{ CVPR'21}$~\cite{edam}  & V2$^\ddag$ & WR38 & 70.9 & 70.6\\
			&AuxSegNet$_\text{ ICCV'21}$~\cite{AuxSeg}  & WR38 & - & 69.0 & 68.6 \\	
			\midrule
			\multirow{11}{*}{$\mathcal{I}$}& \multicolumn{5}{l}{\textit{\textbf{Image-level supervision only.}}}\\
			&IAL$_\text{ IJCV'20}$~\cite{ial}  & V2  & - &  64.3 &  65.4 \\
			&SEAM$_\text{ CVPR'20}$~\cite{seam}  & V3  & WR38 & 64.5 &  65.7 \\
			&BES$_\text{ ECCV'20}$~\cite{bes}  & V2  & R50 & 65.7 &  66.6 \\
			&SC-CAM$_\text{ CVPR'20}$~\cite{sc-cam}  & V2$^\ddag$  & WR38 & 66.1 &  65.9 \\
			&CONTA$_\text{ NeurIPS'20}$~\cite{conta}  & V3  & WR38 & 66.1 &  66.7 \\
			&A$^2$GNN$_\text{ TPAMI'21}$~\cite{a2gnn}  & V2  & WR38 &  66.8 &  67.4 \\
			&VWE$_\text{ IJCAI'2021}$~\cite{vwe}  & V2$^\ddag$  & R50 &  67.2 &  67.3 \\
			&AdvCAM$_\text{ CVPR'21}$~\cite{advcam}  & V2  & R50 & 68.1 &  68.0 \\
			&Kweon \etal$_\text{ ICCV'21}$~\cite{Kweon}  & WR38 & WR38 & 68.4 & 68.2 \\

			\rowcolor{mygray}
			& CLIMS (Ours)  & V2  & R50 & \textbf{69.3} &  \textbf{68.7} \\
			\rowcolor{mygray1}
			&CLIMS (Ours)  & V2$^\ddag$  & R50 & \textbf{70.4} &  \textbf{70.0} \\

			\bottomrule
	\end{tabular}}
\label{tab:segmentation_voc}
\vspace{-10pt}
\end{table}

\begin{figure*}[!htbp]
	\centering
	\includegraphics[width=\linewidth]{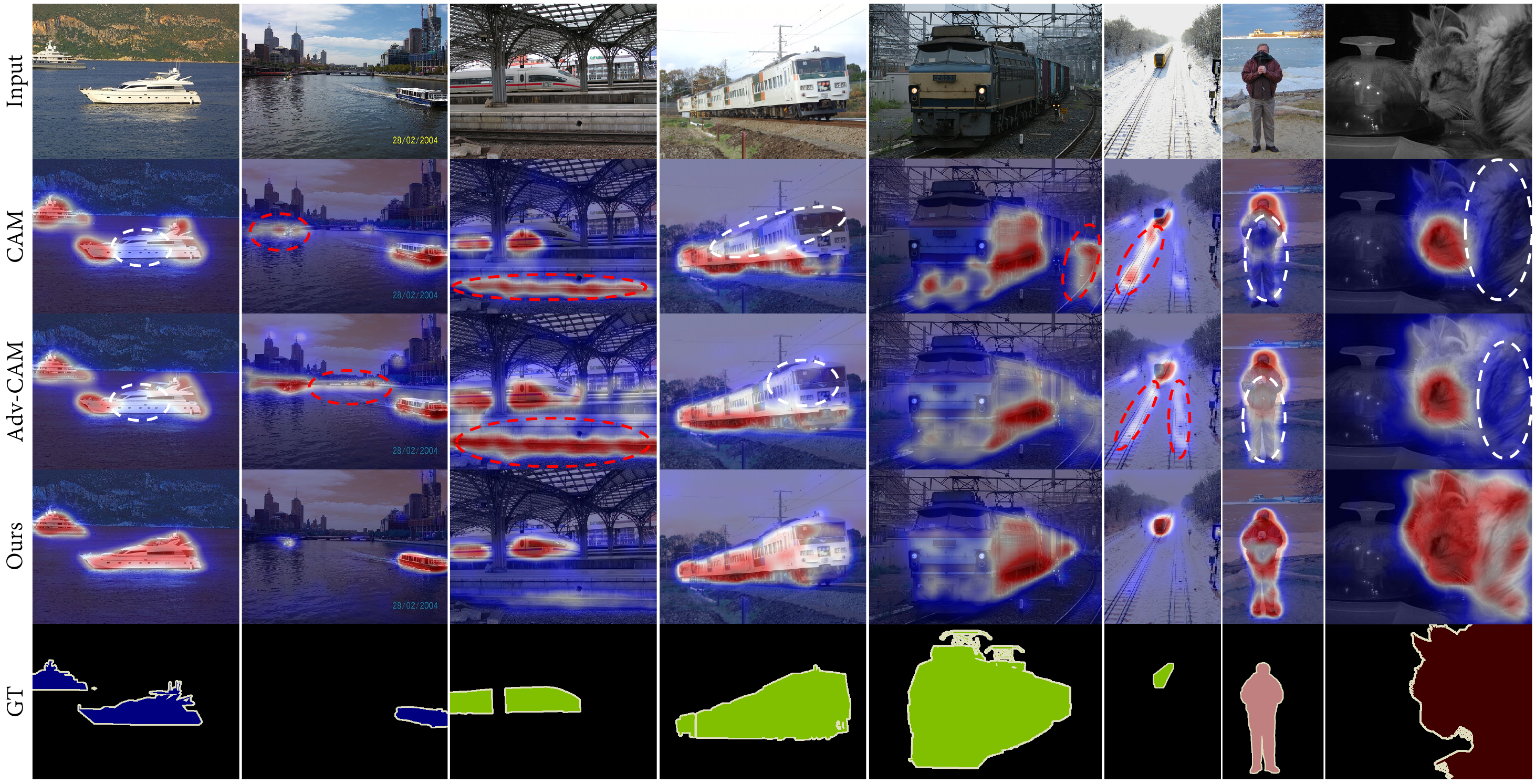}
	\vspace{-20pt}
	\caption{Visualization of the initial CAMs generated by CAM, Adv-CAM, and the proposed CLIMS. White dotted circles illustrate the missed object regions. Red dotted circles illustrate the false activation of class-related background regions, e.g., the river and railroad.}
	\label{fig:vis_campar_sota}
	\vspace{-10pt}
\end{figure*}

\textbf{Segmentation Network.} Given pseudo ground-truth masks, we follow VWE~\cite{vwe}, SC-CAM~\cite{sc-cam} and AdvCAM~\cite{advcam} to adopt DeepLabV2 with ResNet-101~\cite{resnet} as the segmentation network. For experiments on PASCAL VOC2012 dataset, we follow the default setting of deeplab-pytorch toolkit\footnote{https://github.com/kazuto1011/deeplab-pytorch}to train DeepLabV2 with weights pretrained using MS COCO dataset. Besides, we also follow the setting of AdvCAM~\cite{advcam} to train DeepLabV2 with weights pretrained using ImageNet-1K dataset.

\subsection{Results on PASCAL VOC2012 Dataset}
\textbf{Quality of Initial CAMs and Pseudo Labels.} The quality of generated initial CAMs and refined pseudo ground-truth masks on the PASCAL VOC2012 \textit{train} set is compared in Table \ref{tab:refine}. +\textbf{RW} denotes that the initial CAMs are refined by PSA~\cite{affinitynet}. As shown in the table, the initial CAMs generated by our CLIMS reach 56.6\% mIoU, significantly outperforming the state-of-the-art methods, e.g., PuzzleCAM and AdvCAM. This mostly attributes to the designs of three CLIP-based loss functions. When compared to SEAM (WideResNet38~\cite{resnet38} is adopted as the backbone network), our CLIMS (the smaller ResNet-50 is adopted as backbone) achieves better results. Besides, our pseudo labels refined with PSA achieve 70.0\% mIoU, which is 5.3\% and 2.0\% higher than that of PuzzleCAM and AdvCAM, respectively.

\begin{table}[h]
	\centering
	\caption{Comparison of the quality of initial CAMs on PASCAL VOC2012 dataset using different combinations of loss functions. The mIoU(\%) results are reported on the \textit{train} set. $\mathcal{L}_{CLS}$ indicates that we use a pretrained classifier, i.e., VGG-16, to replace our text-driven evaluator for comparsion.}
		\vspace{-10pt}
	\resizebox{\linewidth}{!}{
		\begin{tabular}{c c c c c c}
			\toprule
			$\mathcal{L}_{OTM}$ & $\mathcal{L}_{BTM}$ & $\mathcal{L}_{REG}$ & $\mathcal{L}_{CBS}$ & $\mathcal{L}_{CLS}$ & mIoU(\%) \\
			\midrule
						 & & & & \checkmark & 28.6\\
			\midrule
			\checkmark  &   &   &  & & 37.2 \\
			\checkmark  & \checkmark  &   &  & & 41.3 \\
			\checkmark  & \checkmark  & \checkmark  &  & & 53.1 \\
			\checkmark  & \checkmark  &   & \checkmark & & 45.4 \\
			\rowcolor{mygray}
			\checkmark  & \checkmark  & \checkmark  &  \checkmark & & \textbf{56.6} \\
			
			\bottomrule
	\end{tabular}}
	\label{tab:ablation}
	\vspace{-10pt}
\end{table}

In Fig.~\ref{fig:vis_campar_sota}, we visualize the initial CAMs and compare them with the results of conventional CAM~\cite{zhou2016learning} and recent method Adv-CAM~\cite{advcam}. It is observed that the proposed CLIMS typically activates more complete object contents and less class-related background regions. Specifically, as shown in the first two and last two columns, CAM and Adv-CAM may underestimate the region of boat, person, and cat or falsely activate the region of river. Compared with them, initial CAMs generated by CLIMS are more complete and compact, including more reasonable object regions for next stage of refinement. In the third and sixth columns, lots of false activations of class-related background, i.e., railroad, are available in the results of CAM and Adv-CAM. With the supervision provided by $\mathcal{L}_{CBS}$, our CLIMS can efficiently reduce the false activation of co-occurring background regions and yield complete and compact CAMs.

\textbf{Segmentation Performance.} To further validate the quality of the pseudo ground-truth masks generated by our method, we fully train a segmentation network, i.e., DeepLabV2, on the PASCAL VOC2012 dataset using the generated pseudo labels. The evaluation results are reported in Table \ref{tab:segmentation_voc}. It is observed that, compared to the methods with only image-level supervision, our method achieves the best results. Specifically, DeepLabV2 trained using the mask generated by our CLIMS achieves 69.3\% and 68.7\% mIoU  on \textit{val} and \textit{test} set, respectively. Compared with Adv-CAM, our method outperforms it by 1.2\% and 0.7\% mIoU on \textit{val} and \textit{test} sets, respectively. Compared to methods with additional saliency map (obtained from a fully supervised model), e.g., EPS~\cite{eps} and EDAM~\cite{edam}, our method also achieve competitive performance. 

\begin{table}[!htbp]
	\centering
	\caption{Evaluation results of two specific object categories, i.e., boat and train, on PASCAL VOC2012 dataset with/out the addition of $\mathcal{L}_{CBS}$. The results are reported on the \textit{train} set.}
	\vspace{-10pt}
	\resizebox{\linewidth}{!}{
		\begin{tabular}{l l l l l}
			\toprule
			loss functions. & boat(\%) & train(\%) & mean(\%) \\
			\midrule
			$\mathcal{L}_{OTM}+\mathcal{L}_{BTM}$& 7.1 & 30.7 & 18.9 \\
			\rowcolor{mygray}
			$\mathcal{L}_{OTM}+\mathcal{L}_{BTM} + \mathcal{L}_{CBS}$& \textbf{58.2}$_{+51.1}$  & \textbf{63.9}$_{+33.2}$  & \textbf{61.1}$_{+42.2}$   \\
			\bottomrule
	\end{tabular}}
	\label{tab:effect_of_background}
	\vspace{-8pt}
\end{table}

\begin{figure*}[!htbp]
	\begin{subfigure}{.24\textwidth}
		\centering
		\includegraphics[width=\textwidth]{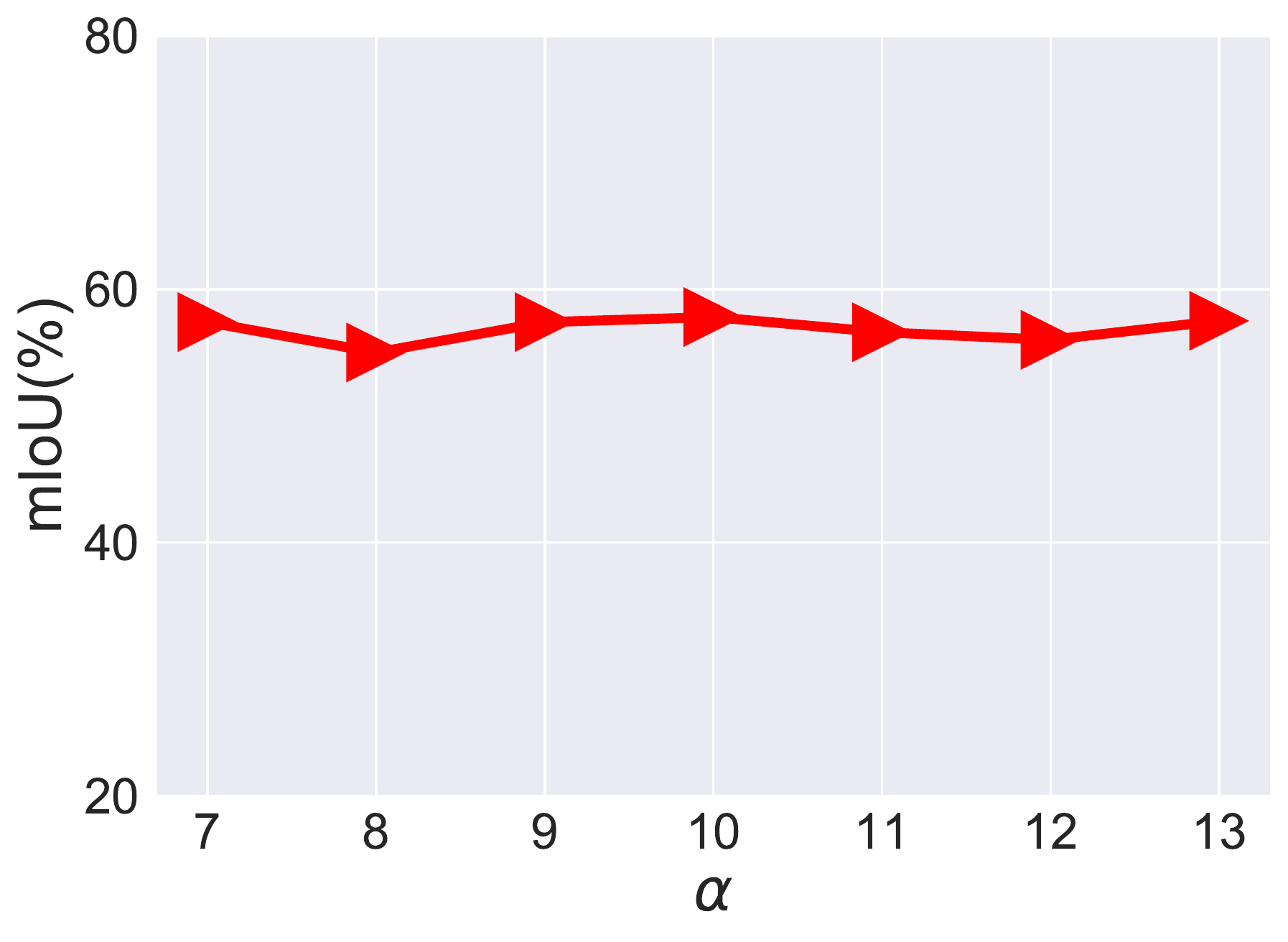}
		\caption{Analysis of hyper-parameters $\alpha$.}
	\end{subfigure}
	\begin{subfigure}{.24\textwidth}
		\centering
		\includegraphics[width=\textwidth]{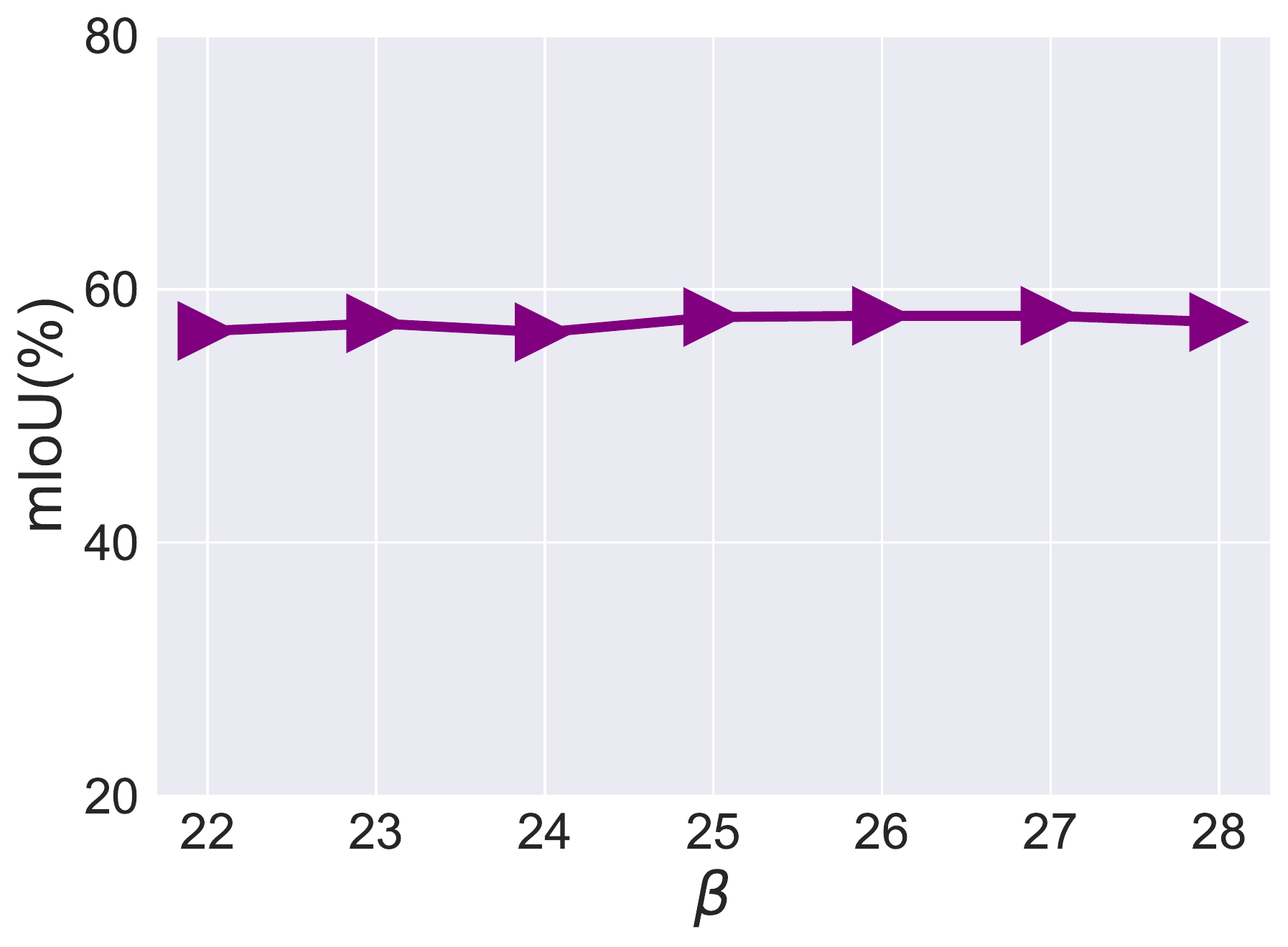}
		\caption{Analysis of hyper-parameters $\beta$.}
	\end{subfigure}
	\begin{subfigure}{.24\textwidth}
		\centering
		\includegraphics[width=\textwidth]{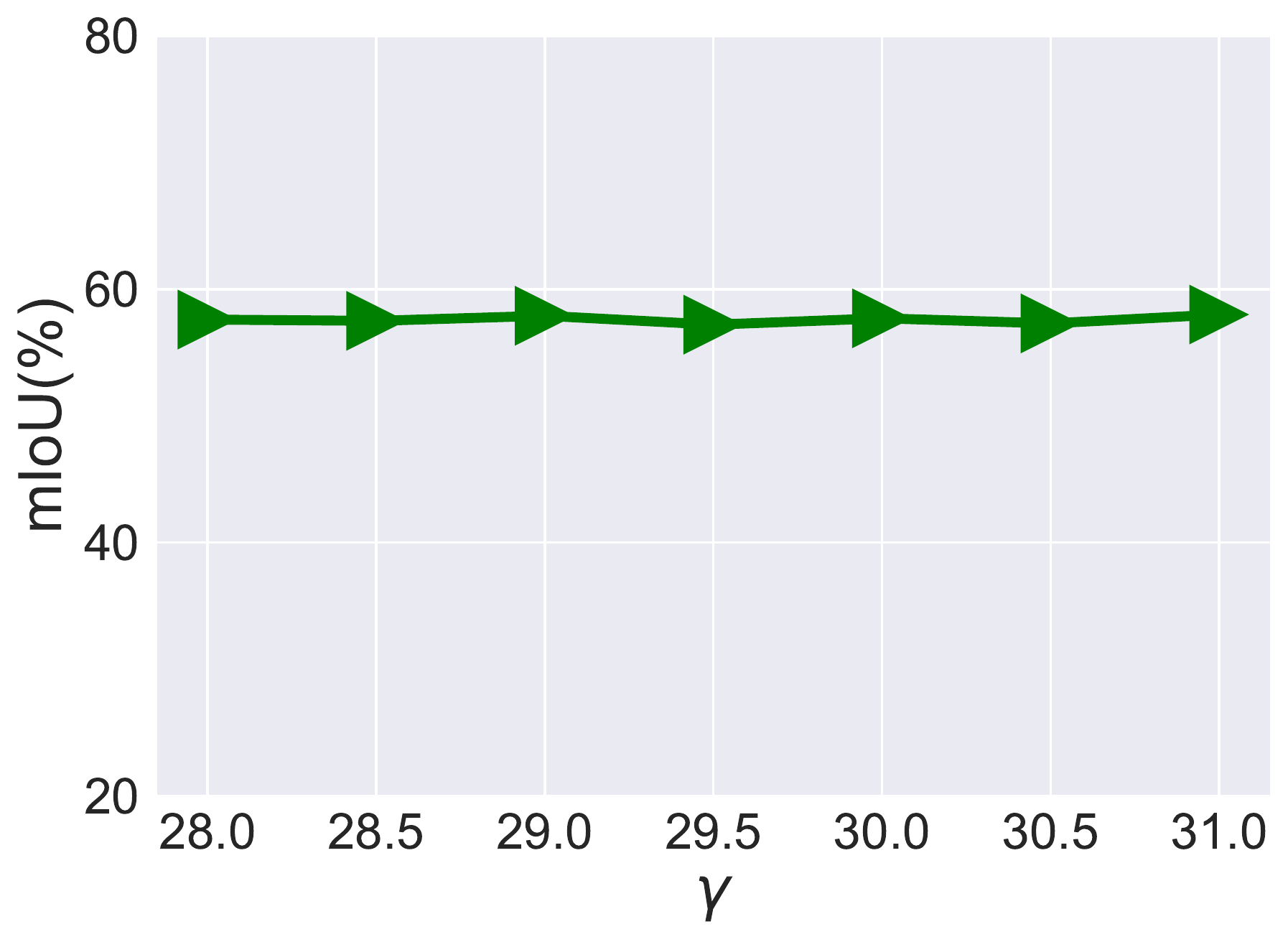}
		\caption{Analysis of hyper-parameters $\gamma$.}
	\end{subfigure}
	\begin{subfigure}{.24\textwidth}
		\centering
		\includegraphics[width=\textwidth]{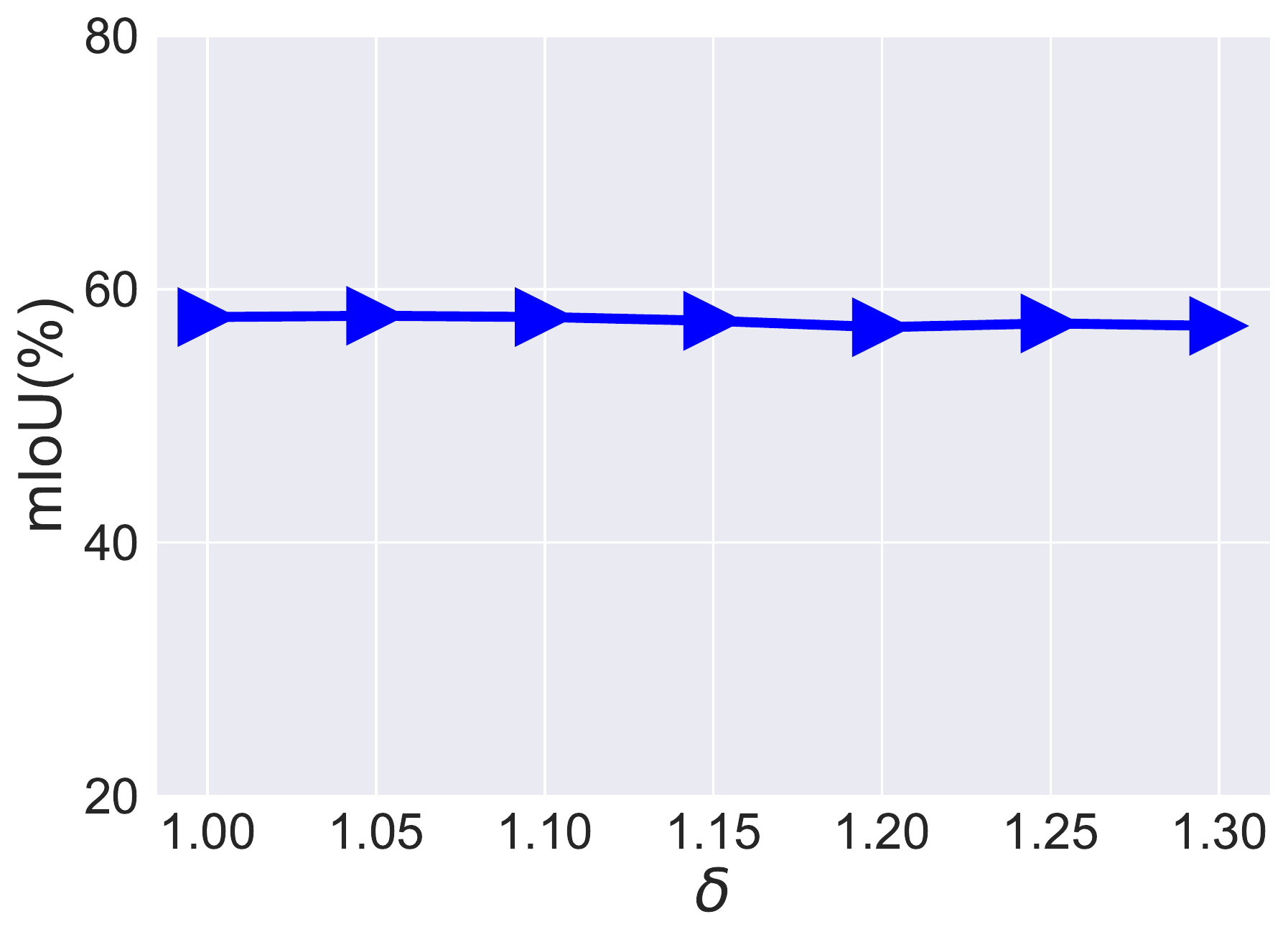}
		\caption{Analysis of hyper-parameters $\delta$.}
	\end{subfigure}
	\vspace{-10pt}
	\caption{Sensitivity analyses of hyper-parameters $\alpha$, $\beta$, $\gamma$, and $\delta$. The mIoU values here are reported on PASCAL VOC2012 \textit{train} set.}
	\label{fig:sensitivity_1}
	\vspace{-10pt}
\end{figure*}
\subsection{Ablation Studies}

\textbf{Impact of Loss Functions.} The proposed CLIMS, as described in Sec.~\ref{sec:method}, is optimized by a combination of four loss functions, i.e., object region and text label matching loss $\mathcal{L}_{OTM}$, background region and text label matching loss $\mathcal{L}_{BTM}$, co-occurring background suppression loss $\mathcal{L}_{CBS}$, and area regularization $\mathcal{L}_{REG}$, which play different roles in guiding backbone network to generate initial CAMs. Here, we perform ablation experiments on the PASCAL VOC2012 dataset to further validate the effectiveness of each loss function. The visual results are shown in Fig.~\ref{fig:ablation} (note that, $+\mathcal{L}$ means that we successively add the loss function). As shown in the second column of Fig.~\ref{fig:ablation}, by only using $\mathcal{L}_{OTM}$, we observe that 1) only the discriminative object parts are activated in the CAMs; 2) class-related backgrounds, e.g., railroad in the second row and water in the third row, are also activated in the initial CAMs. As seen in the third column, an addition of $\mathcal{L}_{BTM}$ significantly increases the size of activated regions, such that more complete object regions are activated. However, $\mathcal{L}_{OTM}+\mathcal{L}_{BTM}$ falsely activate the background regions. As shown in the fourth and fifth column in Fig.~\ref{fig:ablation}, the addition of $\mathcal{L}_{REG}$ and $\mathcal{L}_{CBS}$ helps address the aforementioned issues. We can observe that $\mathcal{L}_{REG}$ efficiently constrains the size of activated regions and $\mathcal{L}_{CBS}$ significantly excludes the class-related background, e.g., railroad, out of the CAMs. Furthermore, as shown in the sixth column, with the refinement of \textbf{RW} (PSA~\cite{affinitynet}), the pseudo masks are closely similar to the ground-truth.


 Table \ref{tab:ablation} shows quantitative comparisons between different combinations of loss functions. As can be seen, CLIMS only obtains 37.2\% mIoU on the \textit{train} set when only $\mathcal{L}_{OTM}$ is used. An addition of $\mathcal{L}_{BTM}$ improves mIoU from 37.2\% to 41.3\%. By ensuring the compactness of initial CAMs, the inclusion of $\mathcal{L}_{REG}$ improves the mIoU from 41.3\% to 53.1\%. As expected, $\mathcal{L}_{CBS}$ can efficiently remove the class-related background regions from the generated CAMs and improve the mIoU by 3.5\% on the \textit{train} set. To validate the effectiveness of CLIP model, we use a pretrained classifer, i.e., VGG-16, to replace text-driven evalutor and compare their performances. It means that only mask-out area of image is fed into the classifier for evaluation. The result is illustrated by $\mathcal{L}_{CLS}$. As shown in Table \ref{tab:ablation}, $\mathcal{L}_{CLS}$ only achieves 28.6\% mIoU, which is much lower than $\mathcal{L}_{OTM}$(37.2\%). This suggest that, without other constraints, supervision from CLIP leads the backbone network to generate better CAMs than a pretrained classifier.

\textbf{Class-related Background.} In our experiments, we only pre-define a set of class-related backgrounds for some object categories, such as train and boat. For example, the background sets of train and boat are $\{$“railroad”, “railway”, “tree”$\}$ and $\{$“river”, “sea”, “lake”$\}$, respectively. To further demonstrate the effectiveness of co-occurring background suppression loss $\mathcal{L}_{CBS}$, we report the IoU of boat and train in Table \ref{tab:effect_of_background} with/out $\mathcal{L}_{CBS}$. As can be seen, an addition of $\mathcal{L}_{CBS}$ can significantly improve the IoU of both boat and train, from 7.1\% to 58.2\% and 30.7\% to 63.9\%, respectively. The results suggest that CLIP can effectively identify those diverse background categories. Based on the embeddings of background regions and text descriptions, $\mathcal{L}_{CBS}$ can effectively exclude these co-concurring backgrounds from the activated regions of foreground objects.

\textbf{Sensitivity Analyses.} There are four hyper-parameters in Eq.~\ref{eq:13}. The sensitivity analyses of these four parameters are performed on PASCAL VOC2012 \textit{train} set and the evaluation results are presented in Fig.~\ref{fig:sensitivity_1}. It is observed that the performances of our approach are stable with the variation of $\alpha$ (from 7 to 13), $\beta$ (from 22 to 28), $\gamma$ (from 28 to 31), and $\delta$ (from 1 to 1.3), i.e., our approach is not sensitive to hyper-parameters. In our experiments, the default value of $\alpha$, $\beta$, $\gamma$, and $\delta$ are 10, 25, 29.5, and 1.15, respectively. (We select big values for the four hyper-parameters to ensure that enough gradients can be backpropagated to the backbone network to avoid gradient vanishing.)



\begin{figure}[h]
	\centering
	\begin{minipage}[t]{0.48\linewidth}
		\centering
		\includegraphics[width=\linewidth]{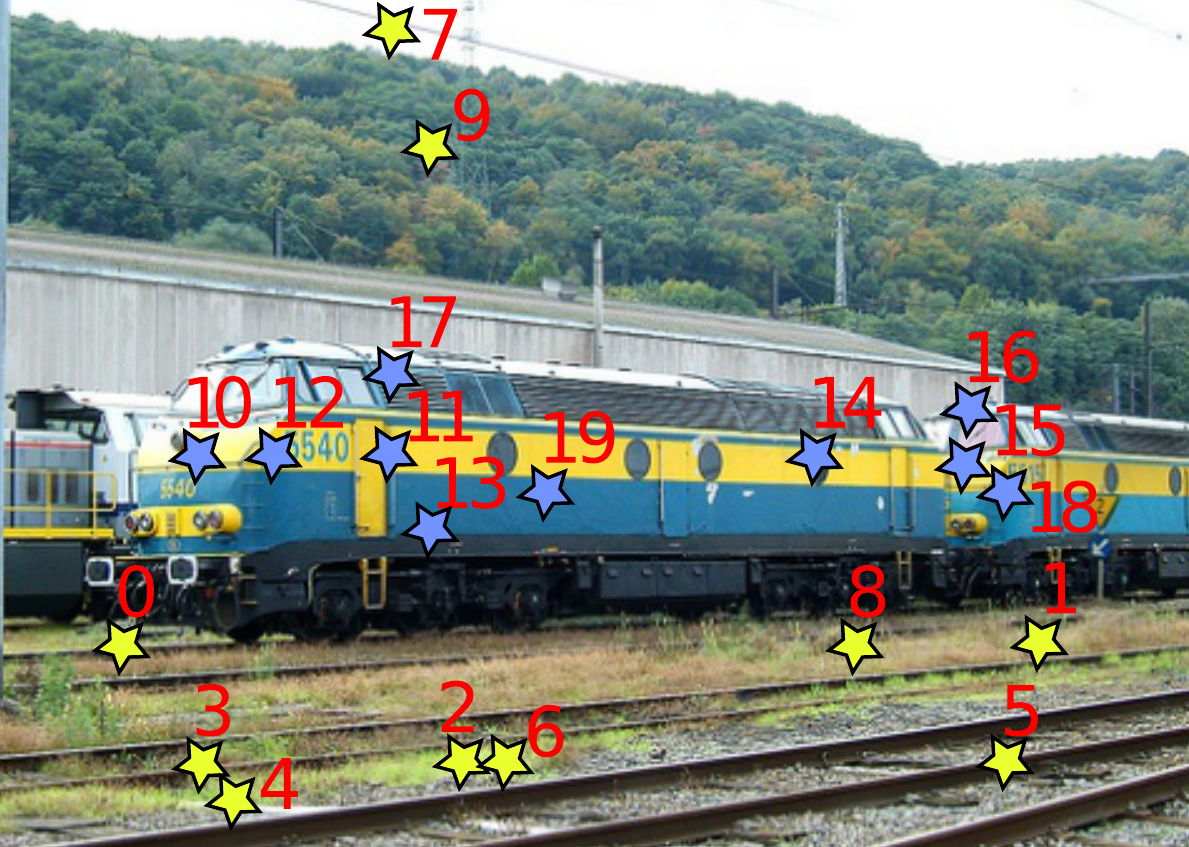}
	\end{minipage}
	\begin{minipage}[t]{0.48\linewidth}
		\centering
		\includegraphics[width=\linewidth]{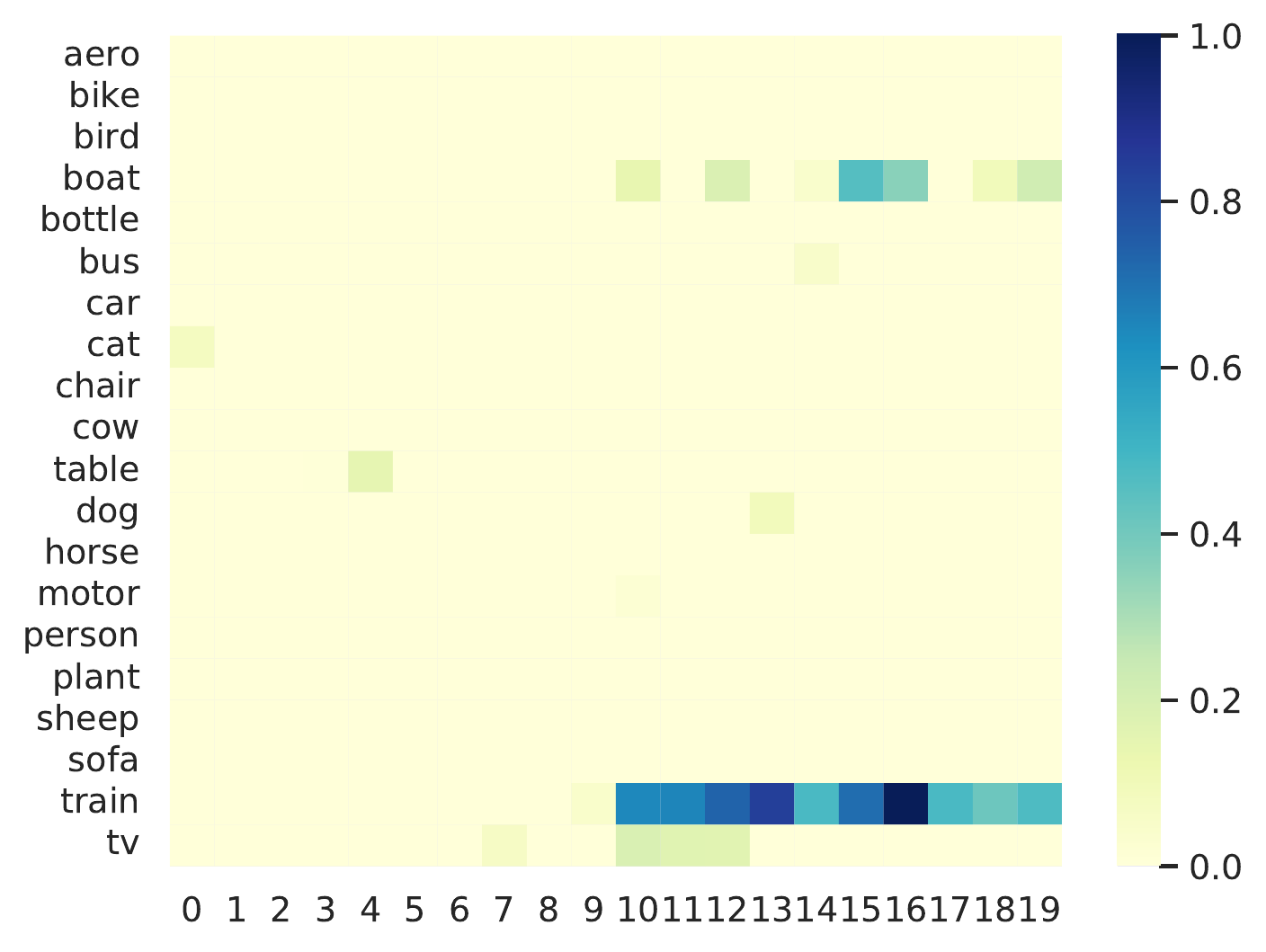}
	\end{minipage}
\vspace{-4pt}
	\caption{Left: the sample image. Yellow and blue stars in the image denote the sample regions. Right: the similarity matrix. The x-axis denotes the feature of sample region and the y-axis denotes the weight vector $\mW_k$ of each class. The $(i,j)$-th element means the cosine similarity between $i$-th class and $j$-th region in the image. Note that, the calculated cosine similarities are truncated and normalized into [0, 1]. Best viewed in color.}
	\label{fig:feature_repr}
	\vspace{-10pt}
\end{figure}

\textbf{Feature Representation Analysis.} Fig.~\ref{fig:feature_repr} depicts the relationship between learned features and the weight vector $\mW_k$. The demo image used to extract representation features is shown in the left of Fig.~\ref{fig:feature_repr}. To investigate the quality of learned features, we sample some features from the pixels belonging to the class of train and background and compute the similarity between weight vector $\mW_k$ and them. The results are shown in the right of Fig.~\ref{fig:feature_repr}. The sample features are denoted in the x-axis, and the weight vector of a class is denoted in the y-axis. It is observed that the extracted features of the blue stars (index from 10 to 19) are highly related to the weight vector $\mW_k$ of train and are unrelated to the other classes. This suggests that the features of train are clustered with $\mW_k$ of train and distribute far away from other classes. The sample regions from the class-related background are represented by yellow stars (index from 0 to 9). Their learned features have a weak relationship with the weight vector of train. This ensures that the class-related background will not be activated in the activation map generation for the class of train.


\section{Conclusion}
This paper proposes a novel Cross Language Image Matching framework, i.e., CLIMS, to introduce natural language supervision for WSSS. The designed four loss functions can efficiently handle the problem of underestimation of complete object contents and unnecessary activation of closely-related background regions.  Extensive experiments conducted on the PASCAL VOC2012 dataset validate the effectiveness of CLIMS. The experimental results reveal that our method generates more complete and compact initial CAMs and refined pseudo ground-truth masks than baseline and state-of-the-art methods.

\clearpage
{\small
\bibliographystyle{ieee_fullname}
\bibliography{egbib}
}

\end{document}